\documentclass[runningheads]{llncs}
\usepackage{graphicx}
\usepackage{booktabs}
\usepackage[utf8]{inputenc}
\usepackage[T1]{fontenc}
\usepackage{bm}
\usepackage{amsmath,amsfonts}
\usepackage{lineno}%
\usepackage{makecell}
\usepackage{scrextend}
\usepackage{float}
\usepackage{hyperref}

\newcommand{\bi}{\begin{itemize}}
\newcommand{\ei}{\end{itemize}}
\newcommand{\ba}{\begin{array}}
\newcommand{\ea}{\end{array}}

\usepackage{xspace}

\newcommand{\bmx}[0]{\begin{bmatrix}}
\newcommand{\emx}[0]{\end{bmatrix}}

\newif\ifarxiv
\arxivfalse 

\begin{document}
\title{Mathematical models for off-ball scoring prediction in basketball}
\titlerunning{Mathematical models for off-ball scoring prediction in basketball}
%

\author{Rikako Kono\inst{1} \and Keisuke Fujii\inst{2,3,4}}
%
\institute{Research School of Physics, Australian National University, Canberra, Australia. \and
Graduate School of Informatics, Nagoya University, Nagoya, Japan. \and
Center for Advanced Intelligence Project, RIKEN, Osaka, Japan.  \and 
PRESTO, Japan Science and Technology Agency, Saitama, Japan. 
\email{rikako.kono@anu.edu.au
fujii@i.nagoya-u.ac.jp}
}
\authorrunning{R. Kono et al.}
%
\vspace{-11pt}

\maketitle              
\begin{abstract}
\vspace{-15pt}
In professional basketball, the accurate prediction of scoring opportunities based on strategic decision-making is crucial for spatial and player evaluations. However, traditional models often face challenges in accounting for the complexities of off-ball movements, which are essential for comprehensive performance evaluations. In this study, we propose two mathematical models to predict off-ball scoring opportunities in basketball, considering pass-to-score and dribble-to-score sequences: the Ball Movement for Off-ball Scoring (BMOS) and the Ball Intercept and Movement for Off-ball Scoring (BIMOS) models. The BMOS model adapts principles from the Off-Ball Scoring Opportunities (OBSO) model, originally designed for soccer, to basketball, whereas the BIMOS model also incorporates the likelihood of interception during ball movements. We evaluated these models using player tracking data from 630 NBA games in the 2015-2016 regular season, demonstrating that the BIMOS model outperforms the BMOS model in terms of team scoring prediction accuracy, while also highlighting its potential for further development. Overall, the BIMOS model provides valuable insights for tactical analysis and player evaluation in basketball.
\ifarxiv
\else
\fi

\vspace{-6pt}
\end{abstract}
\vspace{-20pt}
\section{Introduction}
\vspace{-5pt}
\label{sec:introduction}
    In the highly competitive world of professional basketball, the precise and comprehensive evaluation of player performance is essential for identifying player potential and capability, as well as enabling teams to make well-informed decisions, such as scouting new talent or determining salary allocations. Traditionally, box score statistics, such as points, assists, and rebounds, have been powerful tools for these evaluations \cite{Sampaio03,Page2007UsingBT,Sampaio10,Deshpande2016EstimatingAN}. However, the emergence of tracking technology around 2010 has significantly shifted the way player performance is analyzed. By combining this technology with box score statistics, teams and analysts can glean insights that were previously inaccessible. \par
    
    Most studies using spatiotemporal tracking data have focused on investigating players' movements around the ball, known as ``on-ball movements''. For instance in \cite{yue2014learning}, the authors predicted near-future events such as shooting and passing in basketball based on current game state. Furthermore, basketball shot selections were analyzed using Non-negative matrix factorization, a dimensionality reduction technique \cite{Miller14}, which was extended to consider additional dimensions \cite{Papalexakis18}. 
    Another study introduced a framework to identify on-ball screens \cite{Mcqueen14}, extended to a deep learning model \cite{ai2021novel}, its defense classification \cite{Mcintyre16}, and player prediction for obtaining rebounds \cite{Hojo19}.
    Recently, shot trajectories were analyzed to understand the impact of tight defensive contesting on shot accuracy \cite{daly2020using}. \par

    While on-ball movements certainly deserve attention, movements not directly linked to ball possession, known as ``off-ball movements'', are also pivotal in understanding the complete dynamics of the basketball game
    as is the case for many other team sports  \cite{Lamas2015ModelingTO,Hojo18,teranishi2022evaluation,Lucey14how}.
    Despite the importance of off-ball movements, analyzing the spatiotemporal movements of off-ball players is highly complex due to the multiple potential actions that may interplay with the overall team strategy. Several research attempts have previously been undertaken to handle this complexity. For instance in \cite{wu2022obtracker}, the authors introduced an extensive model for evaluating off-ball movements in shooting and passing situations, which also accounts for player profiles. Another researches analyzed team and player performances using an expected possession value (EPV) model \cite{Cervone14,cervone2016multiresolution}.
    \par

    Although these models offer deep insights, they are generally data-driven approaches that require large quantities of training data and sometimes provide uninterpretable results. This poses a challenge when evaluating continuous changes in player performance from season to season or assessing new players, underscoring the need for models grounded in theory rather than empirical data. Here we propose theory-based basketball models, inspired by the probabilistic mathematical soccer model called OBSO (off-ball scoring opportunities) \cite{OBSO}. They are constructed using physical parameters of players and the ball, providing intuitive and interpretable results. One of our models demonstrates significant improvements in team scoring prediction accuracy, potentially providing valuable insights for tactical analysis and player evaluation in professional basketball. However, this initial model development work also highlights some limitations in our model application. In the following sections, we describe our theoretical framework in Sec. \ref{sec:theory} and experimental methods in Sec. \ref{sec:method}. Next, we present the results and application in Sec. \ref{sec:result} and conclude this paper in Sec. \ref{sec:conclusion}.
    
\vspace{-12pt}
\section{Theoretical Framework}
\vspace{-8pt}
\label{sec:theory}
    The OBSO model \cite{OBSO} estimates the probability that a pass to a target position on the field will lead to a score at a subsequent moment of the possession in soccer, with its applicability limited to the ``pass-to-score'' sequence. However, in compact court invasion sports like basketball, the ``dribble-to-score'' sequence also cannot be overlooked. To address this, we extended the OBSO model to encompass dribbling situations and adapted it to basketball, naming it the Ball Movement for Off-ball Scoring (BMOS) model.
    Additionally, we further developed the Ball Intercept and Movement for Off-ball Scoring (BIMOS) model, which considers
    the possibility of the ball interception during transitions, as factor not expressly covered by the OBSO and BMOS models. \par
    
\begin{figure}[t]
        \vspace{-9pt}
        \centering
        \includegraphics[scale=0.37]{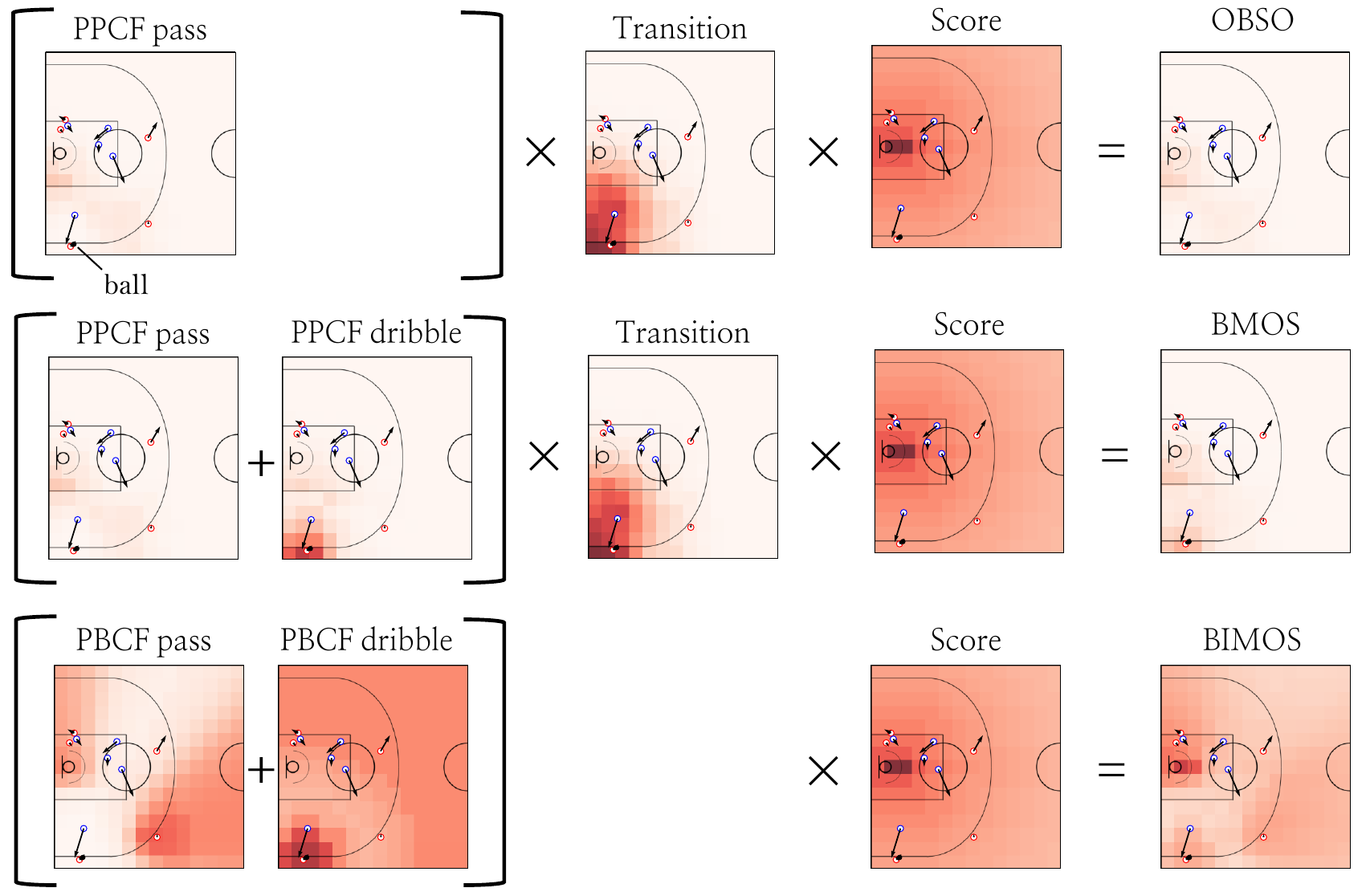}
        \vspace{-9pt}
        \caption{{\bf{Overview of OBSO, BMOS, and BIMOS structures}.}
                Attacking and defending players are represented in red and blue, respectively. An attacker at the bottom left currently possesses the ball. The OBSO and BMOS models consist of three components: spatial occupancy (PPCF), ball delivery probability (Transition Model), and scoring probability (Score Model). The BIMOS model consists of two components: ball occupancy (PBCF) and scoring probability, common to all three models. In the BMOS and BIMOS models, dribbling situations for PPCF and PBCF are incorporated as separate components alongside the pass model.
                }
        \label{fig:bmos_bimos_overview}
        \vspace{-15pt}
    \end{figure}

    Prior to explaining our models in following sections, Fig. \ref{fig:bmos_bimos_overview} provides an overview of the OBSO, BMOS and BIMOS model structures. These models are basically constructed as the multiplication of several step-by-step sub-models. In the OBSO and BMOS models, there are three sub-models: the current spatial occupancy by the attacking team, represented as PPCF (Potential Pitch Control Field); the probability of delivering the ball to a particular location, represented as Transition Model; and the scoring probability, represented as the Score Model. There are two Slightly different sub-models in the BIMOS model: the current ball occupancy by the attacking team, represented as PBCF (Potential Ball Control Field); and the Score Model, which is common to all three models. Briefly, PBCF can be interpreted as a combination of PPCF and the Transition Model, with the added consideration of the ball interception. Moreover, in the BMOS and BIMOS models, we considered dribbling situations for PPCF and PBCF, as separate components alongside the pass model.
    We detail and construct the BMOS and BIMOS models, and demonstrate through comparison that the BIMOS, which explicitly incorporates the interception possibility, represents a more advanced approach.
    \vspace{-10pt}
    \subsection{BMOS Model}
    \vspace{-5pt}
    \label{ssec:bmos_model}
        We segment
        ball movement sequences from its current position to an arbitrary target position $\mathbf{r}\in R\times R$ into three phases according to the OBSO theory \cite{OBSO}: Transition ($\rm{T_r}$), Control ($\rm{C_r}$), and Score ($\rm{S_r}$)
        such that:
        \vspace{-6pt}
        \begin{description}
            \leftskip 10pt
            \item[Transition:] The ball is delivered to point $\mathbf{r}$,
            \item[Control:] The attacking team keeps possession at point $\mathbf{r}$,
            \item[Score:] The attacking team scores from point $\mathbf{r}$.
        \end{description}
        \vspace{-6pt}
        Considering the instantaneous game state \textit{D} which encompasses elements such as players' positions and velocities, the BMOS model calculates the joint probability of these phases and breaks it into a series of conditional probabilities as follows:
        \begin{align}
            \scriptsize
            \text{P}_{\text{BMOS}}=\sum_{\mathbf{r}\in \text{R}\times \text{R}}\text{P}(\text{S}_\mathbf{r}\cap \text{C}_\mathbf{r}\cap \text{T}_\mathbf{r}|\text{D})=\sum_\mathbf{r} \text{P}(\text{S}_\mathbf{r}|\text{C}_\mathbf{r},\text{T}_\mathbf{r},\text{D})\text{P}(\text{C}_\mathbf{r}|\text{T}_\mathbf{r},\text{D})\text{P}(\text{T}_\mathbf{r}|\text{D}).
            \label{eq:obso_original}
        \end{align}
        Here, the conditional probabilities, $\text{P}(\text{S}_\mathbf{r}|\text{C}_\mathbf{r},\text{T}_\mathbf{r},\text{D})$, $\text{P}(\text{C}_\mathbf{r}|\text{T}_\mathbf{r},\text{D})$, and $\text{P}(\text{T}_\mathbf{r}|\text{D})$ are defined as the Score Model, PPCF, and Transition Model, respectively. \\

        \vspace{-8pt}
        \noindent
        \textbf{PPCF.}
            The original PPCF \cite{OBSO} estimates the probability of the attacking team maintaining ball possession at point $\mathbf{r}$ following a successful ball transition, under the condition \textit{D}. In other words, it can be considered as the space occupancy by each team. The PPCF value for player $i$ is numerically calculated as: 
            \begin{align}
                \scriptsize
                \int_{t=\text{T}_{calc}}^{\infty}\text{dPPCF}_i(\mathbf{r},t|\text{D})= 
                \int_{\text{T}_{calc}}^{\infty}\left(1-\sum_k{\text{PPCF}_k(\mathbf{r},t-dt|\text{D})}\right)f_i(\mathbf{r},t|\text{D})\lambda_i dt.
                \label{eq:ppcf_original}
            \end{align}
            Here, $t$ is the time elapsed since the ball starts its movement and $\text{T}_{calc}$ is the expected time for the ball to reach point $\mathbf{r}$. The PPCF value is presumed to be zero when $t<\text{T}_{calc}$. The expression $(1-\sum_k{\text{PPCF}_k})$ denotes the probability that no player has gained possession before $t$. The function $f_i(\mathbf{r},t|\text{D})$ estimates the probability that player $i$ will reach point $\mathbf{r}$ before $t$, while the parameter $\lambda_i$ reflects the player's ball-handling ability. \par
            
            In the BMOS model, we separate the PPCF into pass and dribble components by adjusting ball speeds and the attackers involved in the calculations. Detailed explanations and our modifications from the original theory to $\text{T}_{calc}$, $f_i(\mathbf{r}|\text{D})$, and $\lambda_i$ are presented in the following paragraphs: Time of Flight, Time to Intercept, and Control Rate, respectively. \\

            \vspace{-7pt}
            \noindent
            \textit{Time of Flight.}
                The original PPCF determines $\text{T}_{calc}$ by considering both parabolic trajectories and aerodynamic drag, selecting the most suitable trajectory that best matches the anticipated arrival time of the nearest attacking player \cite{OBSO}. Our adaptation to basketball simplifies this approach by calculating $\text{T}_{calc}$ under the assumption that the ball moves at a constant speed $v_{ball}$ as a function of travel distance, based on the type of ball movement (pass or dribble). \\
                
            \vspace{-7pt}
            \noindent
            \textit{Time to Intercept.}
                To establish $f_i(\mathbf{r},t|\text{D})$, the expected time $\tau_{exp}$ taken for player $i$ to reach point $\mathbf{r}$ is calculated. In the original PPCF, this calculation simplifies a game scenario by assuming that players move directly to point $\mathbf{r}$ at uniform acceleration $\mathbf{a}$ with initial velocity $\mathbf{v}_{ini}$, aligned in the same direction with a realistic upper velocity limit $v_{max}$ \cite{OBSO}. We modified this calculation by incorporating player reaction time, which is the duration taken for a player's cognitive processing to change direction,
                using the approach implemented in \cite{LaurieOnTracking} (a different PPCF implementation). We set the ball possessor’s reaction time to zero and separately accounted for the reaction times of attackers and defenders. Although $\tau_{exp}$ calculations do not include more intricate factors that may influence the true time taken $\tau_{true}$, such as tracking data inaccuracies and tactical decision making, the distribution of residuals $\tau_{true}-\tau_{exp}$ can offer insights into these factors because its normalized cumulative distribution as a function of $t-\tau_{exp}$ yields the probability that a player expected to arrive in $\tau_{exp}$ will actually arrive before $t$. Therefore, its fitted function can be expressed as $f_i(\mathbf{r},t|\text{D})$. In this study, we set $t$ to the expected ball arrival time $\text{T}_{calc}$ to adapt to more realistic basketball situations such that players often catch the ball before it hits the floor. As described in Sec. \ref{ssec:parameter_estimation}, we used a cumulative skewed Cauchy distribution, whereas the original PPCF uses a cumulative logistic function. \\

            \vspace{-7pt}
            \noindent
            \textit{Control Rate.}     
                The parameter $\lambda_i=\lambda$ is defined as the inverse of the time required for player $i$ to gain control of the ball upon reaching point $\mathbf{r}$ \cite{OBSO}. Therefore, higher $\lambda_i$ values indicate a player's quicker ball control abilities. Attackers seek complete control to either lead a subsequent shot or maintain possession, whereas defenders are satisfied with diverting the ball's direction. To differentiate between these objectives, a scaling parameter $\kappa$ $(\geq1)$ is employed for defenders as $\lambda_i=\kappa\lambda$ \cite{OBSO}. \\

        \vspace{-8pt}
        \noindent
        \textbf{Score Model.}
            This model estimates the scoring probability from point $\mathbf{r}$ following a successful transition and control at point $\mathbf{r}$ under the condition \textit{D}. Since the successful transition $\text{T}_{\mathbf{r}}$ and control $\text{C}_{\mathbf{r}}$ conditions are already satisfied during the shooting phase, the scoring probability $\text{S}_d$ can be defined as a function of the distance to the goal $\mathbf{r}_{goal}$. For simplicity, the condition $\textit{D}$ is not directly factored into the model  \cite{OBSO}. The original Score model employs the adjustment parameter $\beta$ to reflect the players' decision-making influence (i.e., $\text{P}(\text{S}_\mathbf{r}, \beta) = \left[\text{S}_d(|\mathbf{r}-\mathbf{r}_{goal}|)\right]^\beta)$ as a tendency to avoid shots under heavy defensive pressure. Although this is a plausible assumption, 
            our scoring probability is obtained by fitting the model to a game dataset (see Sec. \ref{ssec:model_construction}), and further adjustments based on the BMOS value may lead to overfitting and reduce interpretability. Therefore, we have excluded this factor. \\
            
        \vspace{-8pt}
        \noindent
        \textbf{Transition Model.}
            This model estimates the probability of the ball transitioning to point $\mathbf{r}$ under the condition \textit{D} \cite{OBSO}. Our model is constructed using the displacement ($\Delta x$, $\Delta y$) distribution of passes and dribbles, specifically those leading directly to either shots or turnovers. Although the OBSO attempts to incorporate the decision-making factor (i.e., $\left[\sum_{k\in att}\text{PPCF}_k\right]^\alpha$), as the ball possessor is reluctant to deliver the ball to a lower-controlled position, we also omitted this factor for the same reason as in the Score Model; to avoid overfitting and maintain the Transition Model's independence for enhanced interpretability. Since the resulting distribution does not follow the two-dimensional normal distribution adopted in the OBSO, we used a histogram-based distribution with Gaussian filtering (see Sec. \ref{ssec:model_construction}). Note that the Transition phase in the OBSO and BMOS disregards the possibility of ball interception, and the transition probability should be 100\%. Thus, this model can be interpreted as the location where the ball possessor is inclined to deliver the ball. 
    \vspace{-9pt}
    \subsection{BIMOS Model}
    \vspace{-3pt}
    \label{ssec:bimos_model}
        One important limitation in the OBSO and BMOS models is that they do not explicitly account for the possibility that the ball is intercepted en route to target position $\mathbf{r}$. Although this aspect was partially addressed in a preceding study \cite{Spearman2017PhysicsBasedMO}, it is assumed that all relevant players head to point $\mathbf{r}$ rather than selecting alternate paths that may offer a higher chance for ball interception.
        However, this assumption does not always hold in basketball and potentially oversimplifies game scenarios. To address it, we have rewritten Eq. \ref{eq:obso_original} as:
        \begin{align}
            \text{P}_{\text{BIMOS}}=\sum_\mathbf{r} \text{P}(\text{S}_\mathbf{r}|\text{C}_\mathbf{r}, \text{T}_\mathbf{r},\text{D})\text{P}(\text{C}_\mathbf{r}\cap\text{T}_\mathbf{r}|D).
            \label{eq:bimos}
        \end{align}
        Here, $\text{P}(\text{S}_\mathbf{r}|\text{C}_\mathbf{r}, \text{T}_\mathbf{r},\text{D})$ represents the same Score Model mentioned in Sec. \ref{ssec:bmos_model}, and $\text{P}(\text{C}_\mathbf{r}\cap\text{T}_\mathbf{r}|D)$ is defined as the PBCF, which estimates the probability of the attacking team delivering the ball and maintaining possession at point $\mathbf{r}$, under the condition \textit{D}. In other words, it can be considered as the ball occupancy during the delivering from its current position to target position.
        Unlike the PPCF, the PBCF value is not zero where $t<\text{T}_{calc}$ due to the consideration of ball interception. To account for the possibility that neither the attacker nor the defender can gain possession of the ball, we set an upper integration limit to $\text{T}_{calc}$ rather than integrating to infinity. The function $f^b_i$ is then redefined as the probability that player $i$ reaches point $\mathbf{r}_{mid}(t)=\mathbf{r}_{ball}+\mathbf{v}_{ball}\times t$ before $t$, where $\mathbf{r}_{ball}$ is the starting location of the ball, and $\mathbf{r}-\mathbf{r}_{ball}$ and $\mathbf{v}_{ball}$ are aligned in the same direction. Accordingly, Eq. \ref{eq:ppcf_original}  is then modified as
        \begin{align}
            \scriptsize
            \int_{t=0}^{\text{T}_{calc}}\text{dPBCF}_i(t|\mathbf{r},\text{D})= 
            \int_{0}^{\text{T}_{calc}}\left(1-\sum_k{\text{PBCF}_k(t-dt|\mathbf{r},\text{D})}\right)f^b_i(t|\mathbf{r},\text{D})\lambda_i dt.
            \label{eq:pbcf}
        \end{align}
        Note that we solely modified the target position to be calculated over time;
        the methods for separating pass and dribble components for Time of Flight, Time to Intercept, and Control Rate described in Sec. \ref{ssec:bmos_model} remain unchanged.

\vspace{-8pt}
\section{Methods}
\vspace{-6pt}
\label{sec:method}
    The codes used for this work are available on GitHub at \url{https://github.com/Run-summer/bimos_bmos_basketball}.
    \vspace{-12pt}
    \subsection{Dataset}
    \vspace{-6pt}
    \label{ssec:dataset}
        We used a tracking dataset from STATS SportVU, which covers 630 NBA games from the 2015-2016 regular season. 
        To streamline the analysis, all game sequences were standardized to appear as if they occurred on the left half of the court. Scenes with incomplete data, such as those that were obviously truncated, were excluded from the analysis. We allocated 580 games for training our models and set aside the remaining 50 games for testing. However, due to computational constraints, only 4,000 shot scenes and 600 turnover scenes were used for the likelihood estimation process (see Sec. \ref{ssec:parameter_estimation}).
        
    \vspace{-9pt}
    \subsection{Transition and Score Model Construction}
\begin{figure}[t]
        \vspace{-9pt}
        \centering
        \includegraphics[scale=0.48]{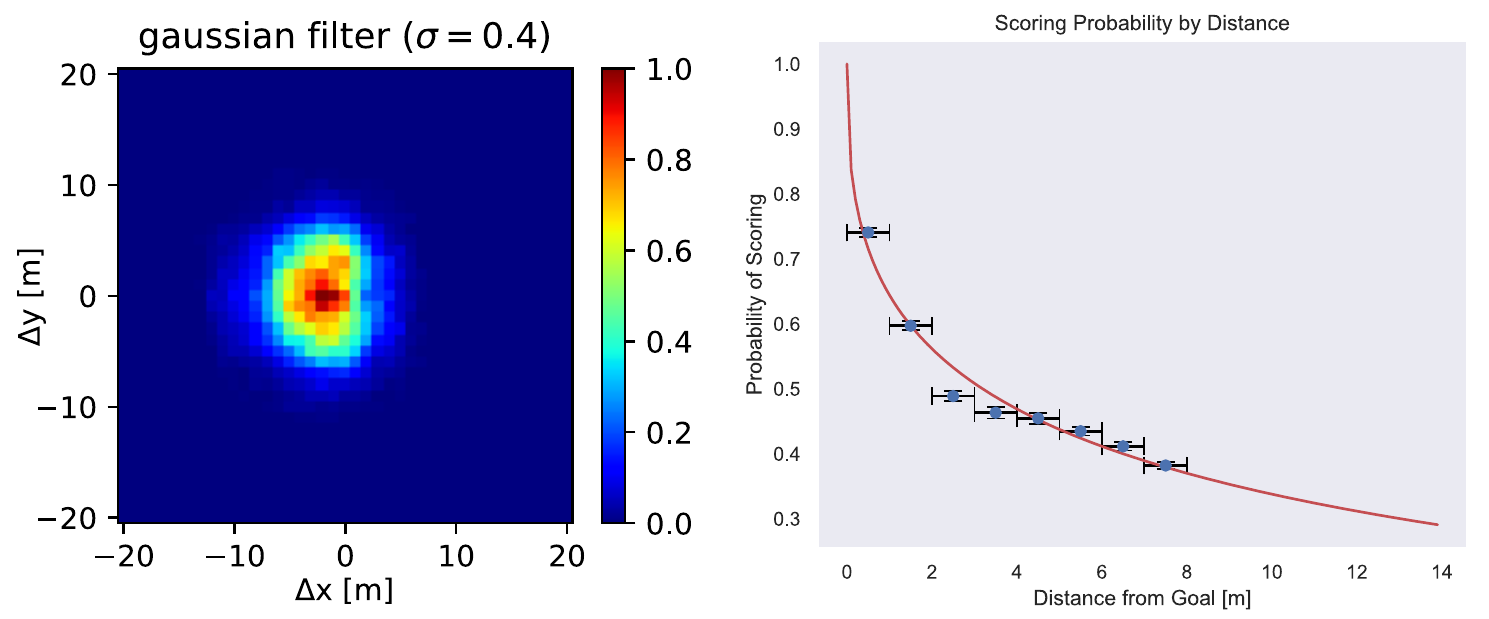}
        \vspace{-10pt}
        \caption{{\bf{Transition Model and Score Model.}}
            Transition Model distribution after applying a Gaussian filter for noise reduction (left). This model captures the tendency of the ball possessor to choose shorter-range ball movement. Scoring probabilities at various distances (right). The exponential function, depicted in red solid line, models the decreasing trend in scoring probability with increasing distance.
            }
        \vspace{-14pt}
        \label{fig:transition_score_model}
    \end{figure}

    \vspace{-3pt}
    \label{ssec:model_construction}
    The Transition Model for the BMOS model was constructed based on the distribution of displacements in pass and dribble transitions leading directly to a shot or turnover. Whereas the OBSO model assumes a two-dimensional normal distribution, our resulting distribution does not follow this assumption, exhibiting a skew towards the negative x-axis, as we only use transitions directly related to shots or turnovers. 
    Therefore, we used a histogram-based distribution with Gaussian filtering ($\sigma=0.4$) to reduce noise and smooth the data (Fig. \ref{fig:transition_score_model} left). \par
    To construct the Score Model, which is common to the BMOS and BIMOS models, scoring probability was calculated based on distance from the goal, considering only distances with more than 2,000 attempts. As expected, the scoring probability decreases with increasing distance, as illustrated in Fig, \ref{fig:transition_score_model} right. To model this trend, an exponential function was employed to fit the data; in contrast, the original study \cite{OBSO} used Gaussian kernel smoothing for a continuous distribution. The fitted function was then extended to cover the entire left half of the basketball court, providing a comprehensive scoring probability map (see Fig. \ref{fig:bmos_bimos_overview} for the resulting Score Model).

    \vspace{-9pt}
    \subsection{PPCF/PBCF Model Parameter Estimation}
    \vspace{-3pt}
    \label{ssec:parameter_estimation}
        Seven parameters need to be determined for the PPCF/PBCF, including the ball velocity $v_{ball}$ for Time of Flight, player maximum velocity $v_{max}$, constant acceleration |\textbf{a}|, reaction times of the attacker and defender for Time to Intercept, and Control Rate parameters $\lambda$ and $\kappa$. Whereas $v_{ball}$ and $v_{max}$ can be directly estimated from the data, the other parameters were fitted using the maximum likelihood estimation technique, as they were challenging to approximate from the data alone. The initial values for these parameters were set according to prior studies \cite{Fujii14,OBSO,Spearman2017PhysicsBasedMO} or our assumptions. \par
    
\begin{figure}[t]
            \vspace{-9pt}
            \centering
            \includegraphics[scale=0.43]{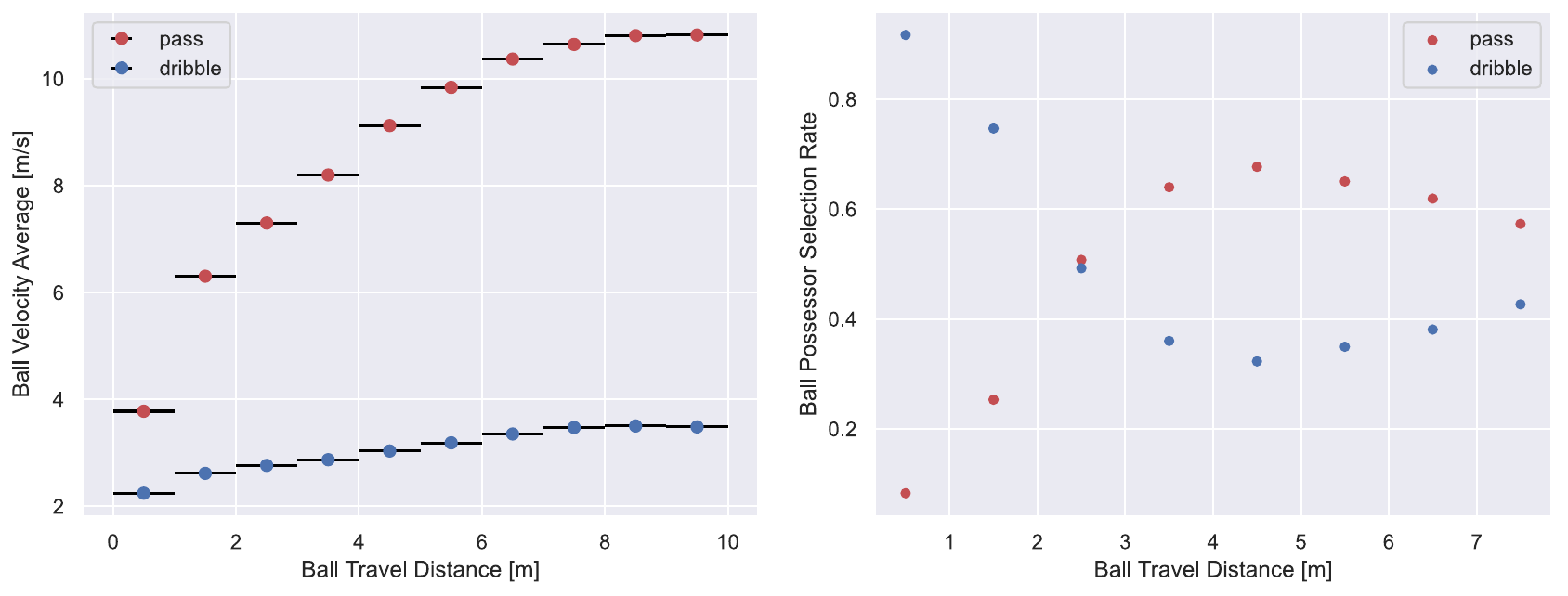}
            \vspace{-12pt}
            \caption{{\bf{Ball Velocity and Ball Handler Selection Rate.}}
                The left figure illustrates how the ball speed varies based on the type of ball movement (pass or dribble) and the distance traveled. For distance exceeding 10 meters, ball speed is assumed constant. The right figure shows the rate at which the ball possessor chooses to pass or dribble based on the travel distance.
                }
            \vspace{-14pt}
            \label{fig:ball_info}
        \end{figure}

        The player maximum velocity $v_{max}$ was set to 5.0 [m/s]. The ball velocity $v_{ball}$ varies between cases of pass and dribble, as well as with distance traveled. Reasonably, ball speed increases with travel distances, and the speed of passes changes more drastically, as shown in Fig. \ref{fig:ball_info} left.
        To combine the pass and dribble PPCF/PBCF, the rate at which the ball possessor decides to pass or dribble was calculated as a function of travel distance, as illustrated in Fig. \ref{fig:ball_info} right. As expected, dribbling was more likely for shorter distances, while passing was the preferred choice for longer distances. \par
\begin{figure}[h]
            \vspace{-14pt}
            \centering
            \includegraphics[scale=0.45]{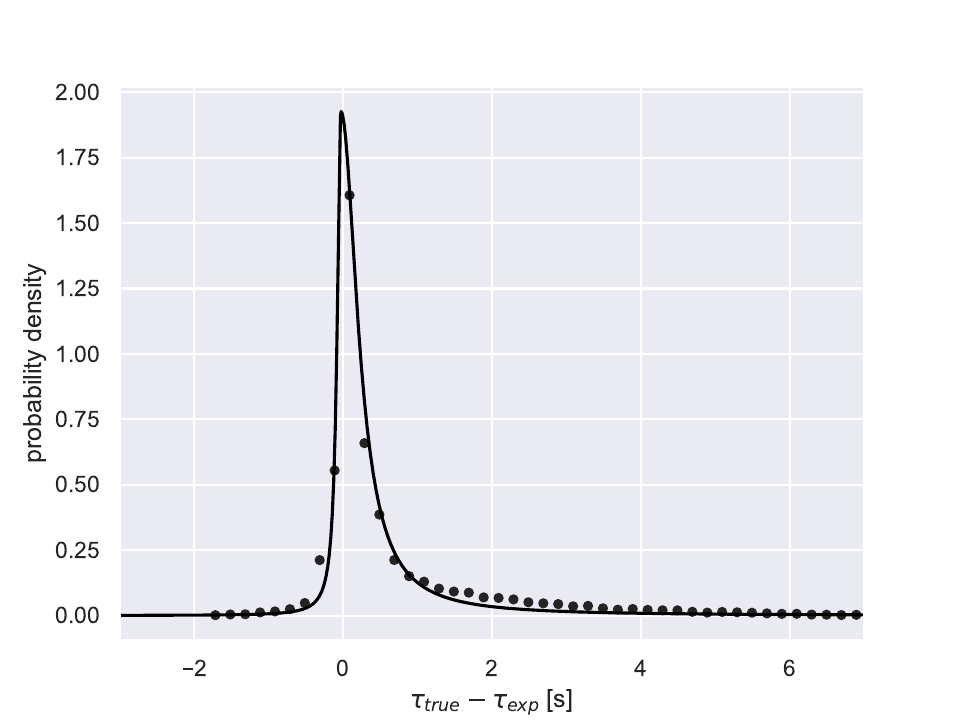}
            \vspace{-12pt}
            \caption{{\bf{$\tau_{true}-\tau_{exp}$ Distribution}.}
                The figure shows the $\tau_{true}-\tau_{exp}$ distribution, highlighting the better fit of an asymmetric Lorentzian function compared to a logistic function. The heavy tail on the positive side suggests that players might not always move at a constant acceleration, resulting in longer than expected times to reach the target position.
                }
            \vspace{-14pt}
            \label{fig:displacement_function}
        \end{figure}

        Initial values of $|\mathbf{a}|$ and $\kappa$ were set to 7.0 [m/s$^2$] and 1.72, respectively, in accordance with the original values \cite{OBSO}. We assumed the time required for a player to gain control of the ball to be much shorter in basketball than soccer, and set $\kappa$ to 30. The attacker and defender reaction times were both set to 0.32 [s] following the outcome of previous study \cite{Fujii14}. The resulting $\tau_{true}-\tau_{exp}$ distribution using these parameters suggests a skewed Cauchy distribution with a heavy tail, where $\tau_{true}-\tau_{exp}>0$
        fits better than a logistic function, as shown in Fig. \ref{fig:displacement_function}.
        This can be attributed to players not actually moving at constant acceleration, taking longer than expected. Consequently, we decided to utilize the accumulated skewed Cauchy distribution as $f_i(\mathbf{r},t|D)$ and $f^b_i(\mathbf{r},t|D)$.
        Using the models constructed above and Eq. \ref{eq:obso_original} or \ref{eq:bimos}, the total BMOS/BIMOS likelihoods are formulated as:
        \vspace{-10pt}
        \begin{equation}
            L(\theta) = \prod_{i=1}^{N} \left[ P(\theta | D)^{y_i} \times (1-P(\theta | D))^{(1-y_i)} \right].
            \nonumber
            \vspace{-6pt}
        \end{equation}
        Here, $N$ represents the number of scenes in the training dataset, $y_i$ indicates whether the $i^{th}$ scene results in a score $y_i~=~1$ or not $y_i~=~0$, and $P(\theta | x_i)$ denotes the BMOS or BIMOS probability at the position of a shot attempt, given condition $D$ and parameters $\theta$. We used values calculated when players passed the ball in pass-to-score sequences, as well as those when players began dribbling in dribble-to-score sequences. Consequently, the maximum likelihood estimation for both the BMOS and BIMOS was performed using the Nelder-Mead method\footnote[1]{The Nelder-Mead algorithm was selected because our model equations are nondifferentiable.}, with boundary conditions set as: $1\le|\textbf{a}|\le8$, $1\le\lambda$, $1\le\kappa$, and $0\le\text{reaction time}\le1$. For the BIMOS, the fitted values were $|\textbf{a}|=7.76$, $\lambda=36.6$, $\kappa=1.02$, attacker reaction time$=0.157$, and defender reaction time$=0.495$. The divergence in reaction time between attackers and defenders can be attributed to differences in tactical understanding, as attackers share tactics in advance, allowing for quicker responses. In the case of the BMOS, the parameter values did not converge under the same maximum likelihood conditions applied to the BIMOS. Therefore, we adopted the estimated parameters of the BIMOS. This is justified because the fitted values obtained through the BIMOS model are both reasonable and realistic, and can therefore be reliably applied to the BMOS model.

\vspace{-12pt}
\section{Results}
\vspace{-8pt}
\label{sec:result}
    \begin{figure}[t]
        \vspace{-8pt}
        \centering
        \includegraphics[scale=0.4]{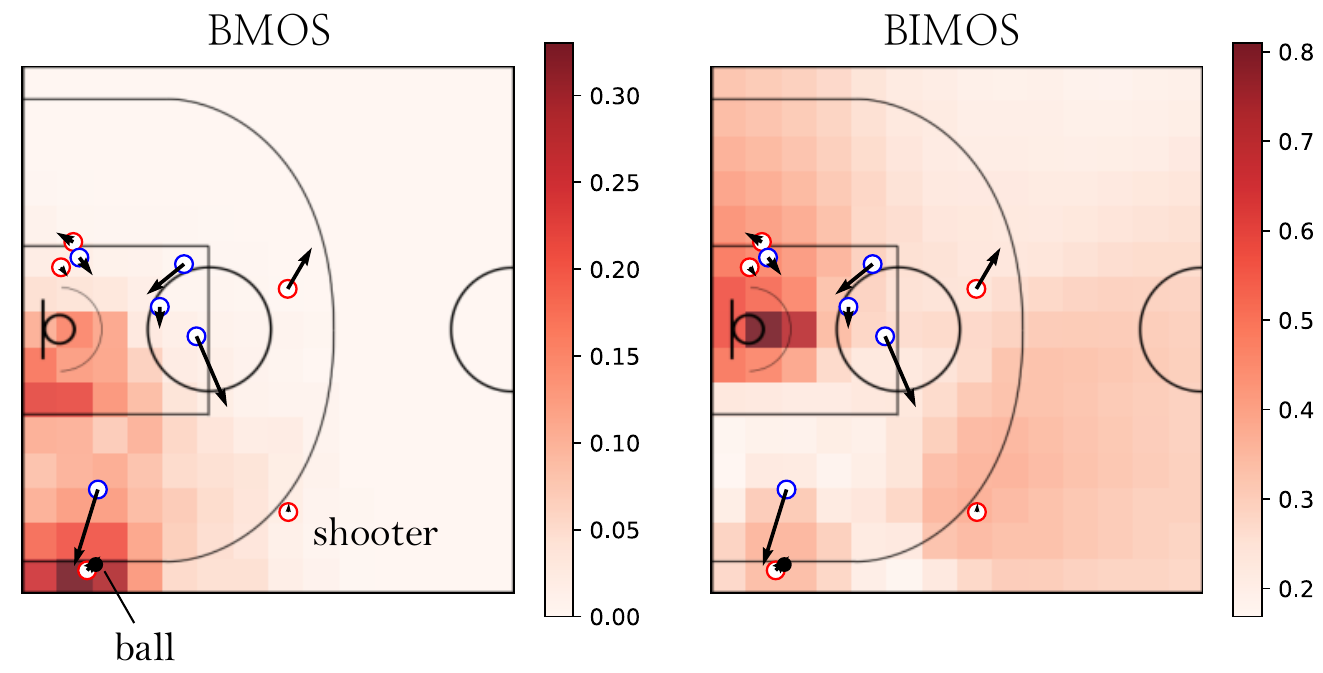}
        \vspace{-12pt}
        \caption{{\bf{BMOS and BIMOS distribution}.}
                The ball possessor at the bottom left passed to a shooter at the bottom middle, resulting in a successful 3-point shot. The BIMOS effectively captures the higher scoring probability in proximity of the shooter.
                }
        \label{fig:observation_comparison}
        \vspace{-12pt}
    \end{figure}
    Fig. \ref{fig:observation_comparison} shows the  BMOS and BIMOS distribution examples, where the ball possessor at the bottom left passed to an attacker at the bottom middle, leading to a successful 3-point shot. The BIMOS captures a reasonable trend, with areas around the current ball possessor, the shooter, and below the goal exhibiting higher scoring probabilities. In contrast, the BMOS failed to describe a higher probability around the shooter. \par
    
\begin{figure}[t]
        \vspace{-0pt}
        \centering
        \includegraphics[scale=0.42]{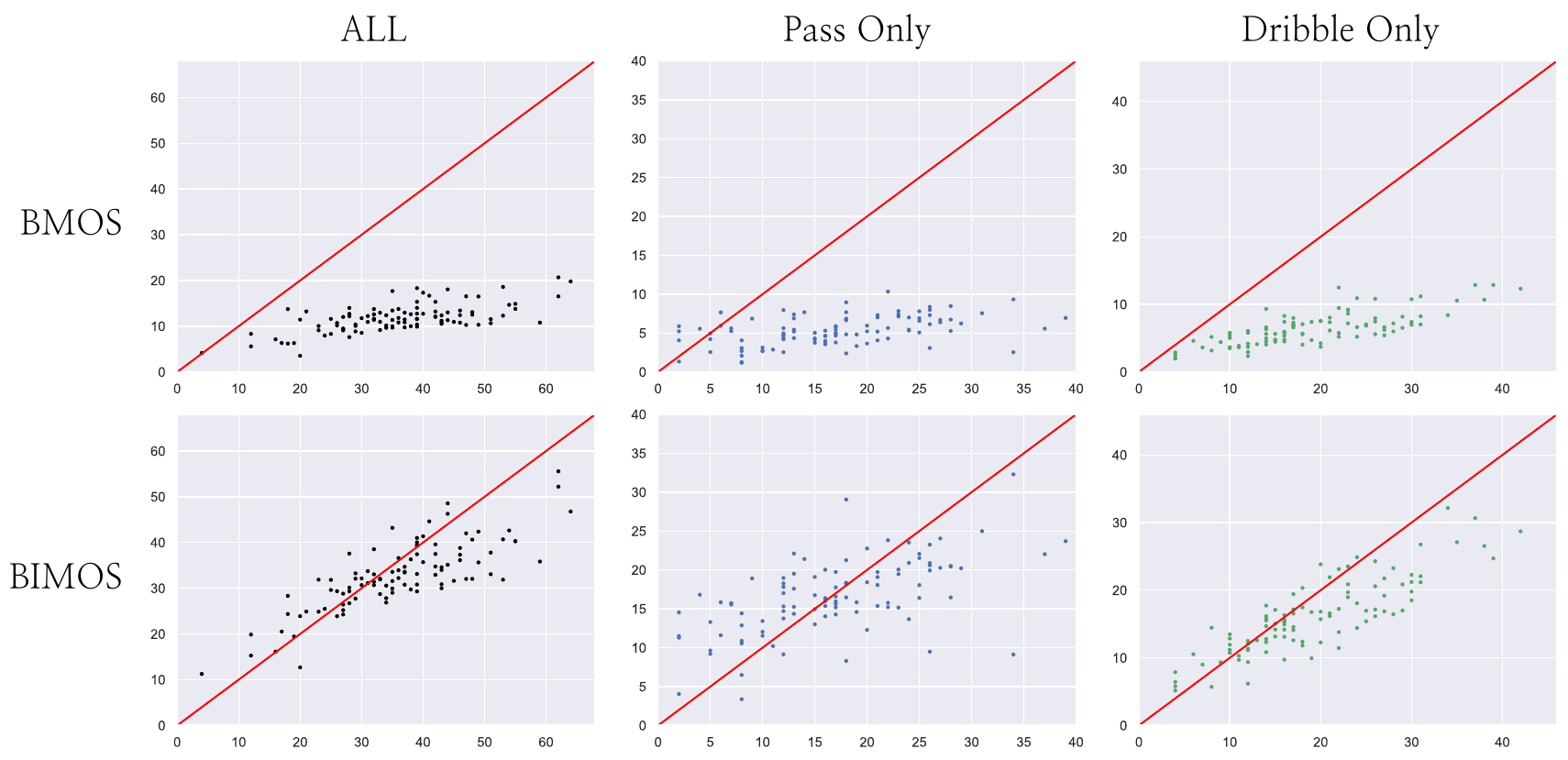}
        \vspace{-15pt}
        \caption{{\bf{Actual vs. Expected score}.}
                Actual total score per game for each team (x-axis) and expected score (y-axis) predicted by the BMOS (top) and BIMOS (bottom) are shown. The middle and right graphs consider only pass-to-score and dribble-to-score sequences, respectively. The red line in each graph represents the identity line. 
                }
        \label{fig:score_comparison}
        \vspace{-16pt}
    \end{figure}

     Next, we quantitatively examined correlations between each team's actual and expected score per game predicted by the BMOS and BIMOS. Fig. \ref{fig:score_comparison} shows the results obtained using 50 test games. Additionally, we divided them into pass-to-score (middle) and dribble-to-score (right) sequences. It should be noted that, since some scenes were not recorded in the original dataset and we also filtered out incomplete data, the actual score does not represent the total score per game, but only the accumulated score from the scenes we used.
    The coefficients of determination $R^2$ for these points with respect to the Expected Score $=$ Actual Score line, depicted in red, are -4.22, -1.75, and -2.05 for the BMOS All, Pass, and Dribble sequences, respectively. In the case of the BIMOS, the respective values are 0.548, 0.576, and 0.547. These results indicate that the BIMOS provides more accurate scoring predictions, especially in pass-to-score sequences. The lower effectiveness in dribble-to-score sequences suggests the need to develop a model specifically designed for dribbling, as both the pass and dribble models were constructed using nearly identical structures for convenience. \\
    
\begin{figure}[t]
        \vspace{-8pt}
        \centering
        \includegraphics[scale=0.5]{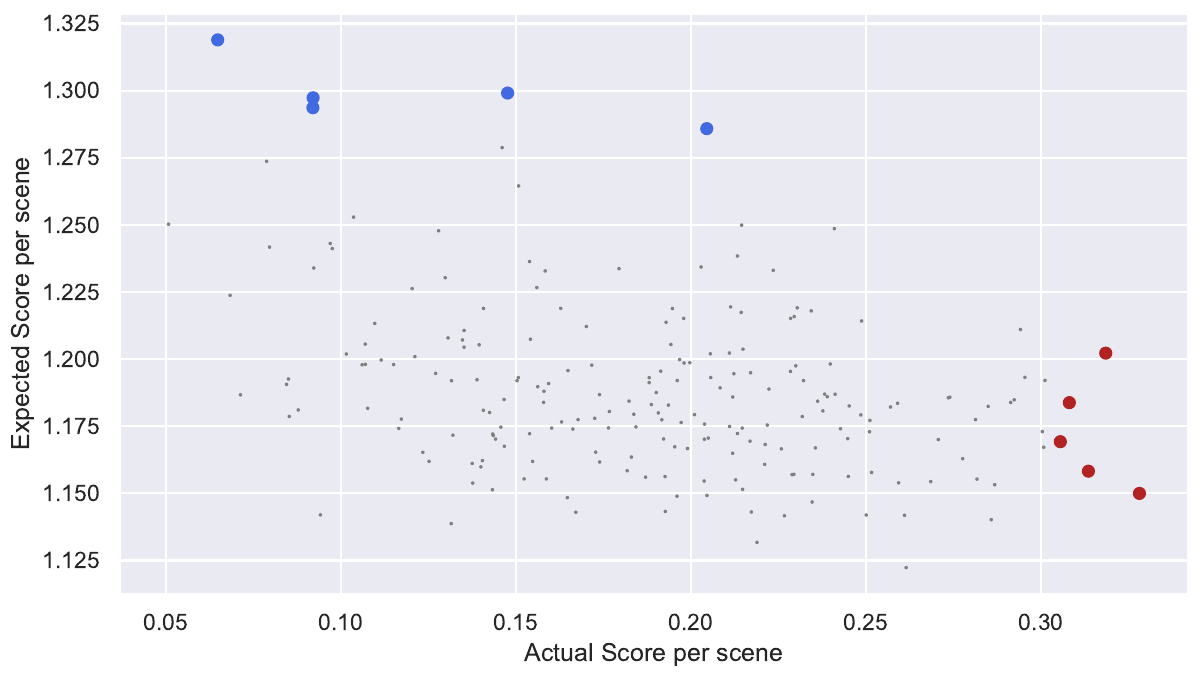}
        \vspace{-12pt}
        \caption{{\bf{Actual vs. Expected score for individuals}.}
                The x-axis represents the actual score per scene for individuals who participated in more than 600 scenes across 630 games, while the y-axis shows the expected scores. The top five players in terms of expected and actual scores are denoted in blue and red, respectively.
                }
        \label{fig:individual_result}
        \vspace{-14pt}
    \end{figure}

    \vspace{-8pt}
    \noindent\textbf{Applications.} Potential applications of the BIMOS model range from tactical reviews to player evaluations. For instance, we show two examples on player evaluations: assessing whether a player maintains position in an area with a high scoring possibility, and determining whether a player creates scoring opportunities by drawing defenders away and moving into other areas. Fig. \ref{fig:individual_result} illustrates the first use case, showing the correlation between expected and actual scores per scene for players who played more than 600 scenes across 630 games. For each scene, we selected the individual's maximum expected score and averaged these values.
    Contrary to our expectation, the overall trend between actual and expected scores is roughly inversely proportional. In terms of expected scores, the top five players played in Center (C) or Power Forward (PF) positions, which are generally responsible for areas around the basket.  Conversely, in terms of actual scores, the top five players played in positions other than Center. 
    This result suggests 
    a limitation of our model: positions around the basket are likely overestimated because our Score Model does not take into account defensive pressure. This limitation should be addressed in future work. Additionally, the second use case---creating scoring opportunities---can be analyzed by computing counterfactuals as described in \cite{umemoto2023evaluation}. Such counterfactual analyses are also expected to enhance tactical reviews by identifying more effective potential tactics. These analyses will also be part of our future work.

\vspace{-12pt}
\section{Conclusions} 
\vspace{-8pt}
\label{sec:conclusion}
    In this study, we extended the OBSO soccer model \cite{OBSO} to the basketball context, designing two mathematical basketball spatial models, the BMOS and BIMOS models, to predict off-ball scoring opportunities considering both pass and dribble situations. The BIMOS model, which also accounts for the possibility of ball interception, significantly improves scoring prediction accuracy compared to the BMOS model, suggesting effective functionality with a limited dataset. \par

    Future work will address current limitations of the BIMOS model, including its lower scoring prediction accuracy in dribble-to-score situations discussed in Sec. \ref{sec:result} and its tendency to overestimate scoring opportunities around the basket described in the results of applications. Additionally, the applications of our model can be further examined in combination with box score statistics and/or computing counterfactuals, as discussed in the results of applications.

\vspace{-8pt}
\section*{Acknowledgments}
\vspace{-8pt}
This study was financially supported by JST PRESTO JPMJPR20CA.
\vspace{-5pt}

\bibliographystyle{splncs04}
\bibliography{reference}


\end{document}